\pdfoutput=1
\documentclass[11pt]{article}

\usepackage[final]{ACL2024}

\usepackage{times}
\usepackage{latexsym}
 \usepackage{enumitem}
\usepackage[normalem]{ulem}

\usepackage[T1]{fontenc}
\usepackage[utf8]{inputenc}

\usepackage{microtype}

\usepackage{inconsolata}
\usepackage{multirow}
\usepackage{booktabs}
\usepackage{graphicx}
\usepackage{todonotes}

\title{Do Large Language Models Speak All Languages Equally?\\ A Comparative Study in Low-Resource Settings}

\author{
Md. Arid Hasan$^1$,   
Prerona Tarannum$^2$,  
Krishno Dey$^1$, 
Imran Razzak$^3$, 
Usman Naseem$^4$\\
$^1$University of New Brunswick, Canada, $^2$Daffodil International University, Bangladesh, \\
$^3$University of New South Wales, Australia, $^4$Macquarie University, Australia \\
\texttt{arid.hasan@unb.ca}
}

\begin{document}
\maketitle
\begin{abstract}

Large language models (LLMs) have garnered significant interest in natural language processing (NLP), particularly their remarkable performance in various downstream tasks in resource-rich languages. Recent studies have highlighted the limitations of LLMs in low-resource languages, primarily focusing on binary classification tasks and giving minimal attention to South Asian languages. These limitations are primarily attributed to constraints such as dataset scarcity, computational costs, and research gaps specific to low-resource languages. To address this gap, we present datasets for sentiment and hate speech tasks by translating from English to Bangla, Hindi, and Urdu, facilitating research in low-resource language processing. Further, we comprehensively examine zero-shot learning using multiple LLMs in English and widely spoken South Asian languages. Our findings indicate that GPT-4 consistently outperforms Llama 2 and Gemini, with English consistently demonstrating superior performance across diverse tasks compared to low-resource languages. Furthermore, our analysis reveals that natural language inference (NLI) exhibits the highest performance among the evaluated tasks, with GPT-4 demonstrating superior capabilities.

\end{abstract}

\section{Introduction}
\label{sec:introduction}

Recent advances in large language models (LLMs) developed significant interest in natural language processing (NLP) across academia and industry. LLMs are known for their language generation capabilities that are trained on billions or trillions of tokens with billions of trainable parameters. Recently researchers have been evaluating LLMs for various NLP downstream tasks, especially question answering \cite{akter2023depth, tan2023can, zhuang2023toolqa}, reasoning \cite{suzgun2022challenging, miao2023selfcheck}, mathematics \cite{lu2023mathvista, rane2023enhancing}, machine translation \cite{xu2023paradigm, lyu2023new}, etc.

Most of the existing works on the evaluation of LLMs are on resource-rich languages such as English. However, the capabilities and performances of LLMs for low-resource languages\footnote{Refers to the scarcity of datasets and other resources rather than limitations in LLM capabilities.} for many NLP downstream tasks are not widely evaluated, leaving a notable gap in the linguistic capabilities of low-resource languages. The most widely spoken yet low-resource languages of South Asia\footnote{https://simple.wikipedia.org/wiki/Languages\_of\_South\_Asia} such as Bangla, Hindi, and Urdu, several researchers are handling the scarcity of datasets and other resources in NLI \cite{aggarwal2022indicxnli}, Sentiment analysis~\cite{hasan2023zero, sun2023sentiment, koto2024zero} and Hate speech detection \cite{khan2021hate, santosh2019hate}. However, the amount of work that uses LLMs is still very few, mainly due to a few constraints such as dataset scarcity, computational costs, and research gaps associated with low-resource languages. These constraints of low-resource languages require more attention, alongside a focus on high-resource languages, to enhance the applicability of LLMs to general-purpose NLP applications.

To fill the aforementioned gap, we comprehensively analyze zero-shot learning using various LLMs in English and low-resource languages. The performance of LLMs shows that GPT-4 provides comparatively better results than Llama 2 and Gemini. Moreover, the English language performs better on different tasks than low-resource languages such as Bangla, Hindi, and Urdu. The Key contributions are as follows:

\begin{itemize}[leftmargin=*] 
    \setlength\itemsep{0em}
    \item To address the limitation of publicly available datasets for low-resource languages, we present datasets for sentiment and hate speech tasks by translating from English to Bangla, Hindi, and Urdu, thereby facilitating research in low-resource language processing.
   
    \item We investigate and analyze the effectiveness of different LLMs across various tasks for both English and low-resource languages such as Bangla, Hindi, and Urdu, which suggest that LLMs perform better when evaluated in English. 

    \item We apply zero-shot prompting using natural language instructions, which describe the task and expected output, enabling constructing a context to generate more appropriate output.
\end{itemize}

%The rest of the paper is organized as follows: Section \ref{sec:related_work} gives a summary of related studies. We describe the detailed zero-shot learning prompting and experimental details in Section \ref{sec:methodology}. Our findings are presented and discussed in the \ref{sec:result_discussion} section. Finally, we concluded in Section \ref{sec:conclusion}.
\section{Related Works}

LLMs are proficient in various NLP tasks and highly generalizable across multiple domains. However, their performance remains significant room for improvement, particularly in low-resource languages such as Bangla, Hindi, and Urdu. Previous study~\cite{robinson2023chatgpt} demonstrates the inability of LLMs such as GPT-4 to perform on low-resource (African) and high-resource languages. However, LLMs perform well in languages (European) that use the same script as English~\cite{holmstrom2023bridging}. %NLI involves determining the logical relationship between pairs of text sequences \cite{conneau2018xnli, kowsher2023contrastive}. LLMs can determine the relationships among text sequences and produce results similar to state-of-the-art techniques \cite{pahwa2023bphigh, gubelmann2023truth}. In contrast, LLMs for sentiment analysis could be a fascinating prospect because we do not have to develop datasets or train models, and it still produces identical results \cite{sun2023sentiment, hasan2023zero}. LLMs have also been studied in hate speech detection \cite{hee2024recent,garcia2023leveraging}. 

NLP research works, and applications for several downstream tasks mainly focus on high-resource languages. Unlike the English language, the advancement of NLP tasks for low-resource languages made it challenging due to several factors described by \cite{alam2021review}. However, there have been some improvements in the last couple of years for Bangla sentiment analysis focusing on resource development \cite{hasan2020sentiment, islam2021sentnob, hasan2023blp} that attained attention from many researchers to concentrate on solving this issue. %However, researchers are focusing on generalizing NLP tasks across the languages. Some of these applications have many limitations for low-resource languages that must be addressed to develop and deploy more generalized universal NLP applications. 
Some of the recent works on NLI \cite{pahwa2023bphigh, gubelmann2023truth}, Sentiment Analysis \cite{xing2024designing, zhang2023enhancing, zhang2023instruct}, and Hate Speech Detection \cite{hee2024recent,garcia2023leveraging} that utilize LLM are mainly carried out in English languages. Moreover, these works opened up the prospects of exploring LLMs for downstream tasks of low-resource languages.

There are few attempts from researchers across different languages to utilize LLM for low-resource languages \cite{hasan2023zero, kabir2023benllmeval,koto2024zero,kumar2021sentiment} that show LLMs can achieve similar results to traditional machine learning techniques and transformer-based models. However, existing multilingual benchmarks such as  BUFFET \cite{asai2023buffet}, XTREME \cite{hu2020xtreme}, XTREME-R \cite{ruder2021xtreme}, MEGA \cite{ahuja2023mega}, and MEGAVERSE \cite{ahuja2023megaverse} do not address all four South Asian low-resource languages we are considering in our study.  Moreover, BUFFET is limited to binary classification tasks and uses few-shot learning and instruction fine-tuning of smaller LLMs (such as mT5, mT0) and ChatGPT. At the same time, we focus on multi-class classification and use zero-shot learning with SOTA LLMs. The performance of LLMs is not balanced for all languages \cite{huang2023not, qin2023cross}, and our study uniquely focuses on comparing resource-rich (English) and low-resource (Bangla, Hindi, and Urdu) languages using SOTA LLMs.

Previous studies have highlighted LLM limitations in low-resource languages, particularly in binary classification, with minimal focus on South Asian languages. These constraints include dataset scarcity, high computational costs, and specific research gaps. To address these challenges, we concentrate on South Asian languages like Bangla, Urdu, and Hindi. We provide datasets for sentiment and hate speech tasks by translating from English. We explore zero-shot learning techniques across English and South Asian languages, thus expanding LLM applications in low-resource settings.

% For low-resource languages, there are significant research gaps in comparison with English. The literature on low-resource languages mainly focused on traditional deep learning and fine-tuning small language models. At the same time, large-scale development has been imposed for resource-rich languages like English. The works that use LLMs to solve downstream tasks in low-resource language are very limited, and the capabilities of LLMs have not been explored properly. To address these issues in this work, we aim to comprehensively use LLMs across several tasks for several low-resource languages such as Bangla, Urdu, and Hindi. 
\section{Methodology}
\label{sec:methodology}

% \subsection{LLMs}

We focused on both open- and closed-source LLMs. We choose three LLMs that are GPT-4~\cite{openai2023gpt}, Llama 2~\cite{touvron2023llama}, and Gemini Pro~\cite{team2023gemini}. We select the LLMs based on their performances, parameter sizes, and capabilities. To conduct our experiments, we used the XNLI dataset~\cite{conneau2018xnli} for the NLI task, the official test of SemEval-2017 task 4~\cite{SemEval:2017:task4} for the sentiment task, and the dataset described in \cite{hateoffensive} for hate speech task. We provide the details of the dataset used and the detailed data preprocessing and evaluation metrics in Appendix \ref{sec:appendix_b}.

% We focused on both open- and closed-source LLMs. We choose three LLMs that are GPT-4~\cite{openai2023gpt}, Llama 2~\cite{touvron2023llama}, and Gemini Pro~\cite{team2023gemini}. We select the LLMs based on their performances, parameter sizes, and capabilities.

% We first discuss the details of prompt techniques and then the LLMs. To conduct our experiments, we used the XNLI dataset\cite{conneau2018xnli} for the NLI task, the official test of SemEval-2017 task 4\cite{SemEval:2017:task4} for the sentiment task, and the dataset described in \cite{hateoffensive} for hate speech task. We provide the details of the dataset used and the detailed data preprocessing and evaluation metrics in Appendix \ref{sec:appendix_b}.

\noindent\textbf{Prompt Approach: }
%\subsection{Prompt Approach}
The performance of LLMs varies depending on the prompt content. Designing a good prompt is a complex and iterative process that requires substantial effort due to the unknown representation of information within the LLM. 
In this study, we applied zero-shot prompting by using natural language instructions. The instructions contain the task description and expected output, which enables the construction of a context to generate more appropriate output. We keep the same prompt for each task across the LLMs. Further, we added role information into the prompt for the GPT-4 model as GPT-4 can take the role information and perform accordingly. We also provide a safety setting for the Gemini model to avoid blocking harmful content. See Appendix \ref{sec:appendix_a} for details.
% of the prompts and safety settings.

% \subsection{LLMs}

% We focused on both open- and closed-source LLMs. We choose three LLMs that are GPT-4~\cite{openai2023gpt}, Llama 2~\cite{touvron2023llama}, and Gemini Pro~\cite{team2023gemini}. We select the LLMs based on their performances, parameter sizes, and capabilities.

%\subsection{Experimental Settings}

%We discuss the detailed data preprocessing and evaluation metrics in Appendix \ref{sec:appendix_b}.

%\subsubsection{Data}
%\input{short_paper/data}

\section{Results and Discussion}
\label{sec:result_discussion}
%This section presents the results and discussion of our experiments and findings.

%\subsection{Results}
%\label{ssec:result}
% This section presents and discusses the performances of different LLMs for English vs low-resource languages. Further, we also discuss the performances of our experiments using different LLMs for different tasks in this section. Table \ref{tab:performance_all} represents the performances of NLI, sentiment, and hate speech tasks.

% \subsection{English \textit{{vs}} Low-resource Languages}

\textbf{English \textit{{vs}} Low-resource Languages:} Our experiments show that all the LLMs consistently provide superior performances for English languages in all tasks except the performances of Gemini in the sentiment task ( Table \ref{tab:performance_all}). In the NLI task, the performance of GPT-4 in English is $18.04$\%, $17.38$\%, and $22.81$\% better than the Bangla, Hindi, and Urdu languages respectively (see Table \ref{tab:performance_all}). Although Hindi performs better than Bangla and Urdu, there is still a massive performance gap compared to English. 
Besides, Llama 2 performance in English is $32.52$\%, $31.28$\%, and $29.94$\% higher compared with Bangla, Hindi, and Urdu respectively. The difference between English and other languages is $\sim$$70$\% from their original performance. Although the performance differences of Gemini between English and other languages are comparatively lower than GPT-4 and Llama 2, English is accomplishing approximately $13$\% better on average than Bangla, Hindi, and Urdu.

\begin{table}[!h]
\centering
\resizebox{0.85\linewidth}{!}{
\begin{tabular}{llrrrr}
\toprule
\multicolumn{1}{c}{\textbf{Model}} & \multicolumn{1}{c}{\textbf{Lang.}} & \multicolumn{1}{c}{\textbf{Acc.}} & \multicolumn{1}{c}{\textbf{P.}} & \multicolumn{1}{c}{\textbf{R.}} & \multicolumn{1}{c}{\textbf{F1$_{macro}$}}\\ \midrule

\multicolumn{6}{c}{\textbf{NLI Task}} \\ \midrule

\multirow{4}{*}{\textbf{GPT-4}} & \textbf{EN} & 86.73 & 86.91 & 86.73 & \textbf{86.79} \\ \cline{2-6}
& \textbf{BN} & 68.73 & 75.95 & 68.73 & \textbf{68.75} \\ \cline{2-6}
& \textbf{HI} & 69.31 & 76.26 & 69.31 & \textbf{69.41} \\ \cline{2-6}
& \textbf{UR} & 64.52 & 72.90 & 64.52 & \textbf{63.98} \\ \midrule

\multirow{4}{*}{\textbf{Llama 2}} & \textbf{EN} & 74.47 & 76.27 & 74.47 & 74.82 \\ \cline{2-6}
& \textbf{BN} & 45.66 & 52.74 & 45.66 & 42.30 \\ \cline{2-6}
& \textbf{HI} & 47.29 & 65.68 & 47.29 & 43.54 \\ \cline{2-6}
& \textbf{UR} & 46.39 & 53.68 & 46.39 & 44.88 \\ \midrule

\multirow{4}{*}{\textbf{Gemini}} & \textbf{EN} & 78.40 & 78.06 & 78.40 & 78.12 \\ \cline{2-6}
& \textbf{BN} & 67.24 & 69.32 & 67.24 & 67.16 \\ \cline{2-6}
& \textbf{HI} & 66.48 & 68.67 & 66.48 & 66.50 \\ \cline{2-6}
& \textbf{UR} & 62.14 & 65.38 & 62.14 & 62.01 \\ \midrule

\multicolumn{6}{c}{\textbf{Sentiment Task}} \\ \midrule

\multirow{4}{*}{\textbf{GPT-4}} & \textbf{EN} & 72.64 & 73.05 & 72.64 & \textbf{71.74} \\ \cline{2-6}
& \textbf{BN} & 61.33 & 64.57 & 61.33 & 56.36 \\ \cline{2-6}
& \textbf{HI} & 66.47 & 68.75 & 66.47 & 63.68 \\ \cline{2-6}
& \textbf{UR} & 62.31 & 64.89 & 62.31 & 58.19 \\ \midrule

\multirow{4}{*}{\textbf{Llama 2}} & \textbf{EN} & 55.64 & 66.89 & 55.64 & 53.38 \\ \cline{2-6}
& \textbf{BN} & 45.19 & 60.22 & 45.19 & 40.28 \\ \cline{2-6}
& \textbf{HI} & 48.31 & 63.32 & 48.31 & 43.73 \\ \cline{2-6}
& \textbf{UR} & 47.06 & 61.61 & 47.06 & 42.62 \\ \midrule

\multirow{4}{*}{\textbf{Gemini}} & \textbf{EN} & 64.59 & 67.86 & 64.59 & 64.44 \\ \cline{2-6}
& \textbf{BN} & 65.40 & 66.68 & 65.40 & \textbf{64.93} \\ \cline{2-6}
& \textbf{HI} & 65.87 & 67.14 & 65.87 & \textbf{65.33} \\ \cline{2-6}
& \textbf{UR} & 65.93 & 66.77 & 65.93 & \textbf{65.14} \\
\midrule

\multicolumn{6}{c}{\textbf{Hate Speech Task}} \\ \midrule

\multirow{4}{*}{\textbf{GPT-4}} & \textbf{EN} & 86.81 & 85.52 & 86.81 & \textbf{62.54} \\ \cline{2-6}
& \textbf{BN} & 55.32 & 75.51 & 55.32 & 38.79 \\ \cline{2-6}
& \textbf{HI} & 64.66 & 77.93 & 64.66 & \textbf{44.61} \\ \cline{2-6}
& \textbf{UR} & 54.00 & 75.18 & 54.00 & 38.66 \\ \midrule

\multirow{4}{*}{\textbf{Llama 2}} & \textbf{EN} & 79.32 & 83.93 & 79.32 & 60.04 \\ \cline{2-6}
& \textbf{BN} & 69.92 & 69.12 & 69.92 & \textbf{41.36} \\ \cline{2-6}
& \textbf{HI} & 74.54 & 71.58 & 74.54 & 44.39 \\ \cline{2-6}
& \textbf{UR} & 47.29 & 65.68 & 47.29 & \textbf{43.54} \\ \midrule

\multirow{4}{*}{\textbf{Gemini}} & \textbf{EN} & 58.00 & 77.69 & 58.00 & 49.10 \\ \cline{2-6}
& \textbf{BN} & 30.34 & 70.93 & 30.34 & 30.81 \\ \cline{2-6}
& \textbf{HI} & 32.01 & 72.72 & 32.01 & 33.36 \\ \cline{2-6}
& \textbf{UR} & 28.56 & 70.07 & 28.56 & 28.47 \\

\bottomrule
\end{tabular}
}
\vspace{-0.2cm}
\caption{Performances of all the tasks across the models and languages. \textbf{Bold} indicates the best performances across the languages for each task. Lang.: language, Acc.: accuracy, P.: Precision, R.: Recall, EN: English, BN: Bangla, HI: Hindi, and UR: Urdu}
\label{tab:performance_all}
\vspace{-0.4cm}
\end{table}

For the sentiment task, English is performing nearly on average $13\%$ better than other languages using GPT-4 (see Table~\ref{tab:performance_all}). The performance difference of Llama 2 between English and other languages is $\sim11\%$ on average, and English is consistently doing better than other languages. Despite that, Bangla, Hindi, and Urdu are performing $0.49\%$, $0.89\%$, and $0.60\%$ better than English. The performance of Gemini remains almost the same for all the languages in the sentiment task.
Our hate speech task experiments reveal that the performance of GPT-4 in English is approximately, on average, $22\%$ better than low-resource languages (see Table \ref{tab:performance_all}). Moreover, the performances in English are $\sim17\%$ and $\sim18\%$ better than low-resource languages for Llama 2 and Gemini models.

% \begin{table}[!ht]
% \centering
% \resizebox{\linewidth}{!}{
% \begin{tabular}{llrrrr}
% \toprule
% \multicolumn{1}{c}{\textbf{Model}} & \multicolumn{1}{c}{\textbf{Lang.}} & \multicolumn{1}{c}{\textbf{Acc.}} & \multicolumn{1}{c}{\textbf{P.}} & \multicolumn{1}{c}{\textbf{R.}} & \multicolumn{1}{c}{\textbf{F1$_{macro}$}}\\ \midrule

% \multirow{4}{*}{\textbf{GPT-4}} & \textbf{EN} & 72.64 & 73.05 & 72.64 & \textbf{71.74} \\ \cline{2-6}
% & \textbf{BN} & 61.33 & 64.57 & 61.33 & 56.36 \\ \cline{2-6}
% & \textbf{HI} & 66.47 & 68.75 & 66.47 & 63.68 \\ \cline{2-6}
% & \textbf{UR} & 62.31 & 64.89 & 62.31 & 58.19 \\ \midrule

% \multirow{4}{*}{\textbf{Llama 2}} & \textbf{EN} & 55.64 & 66.89 & 55.64 & 53.38 \\ \cline{2-6}
% & \textbf{BN} & 45.19 & 60.22 & 45.19 & 40.28 \\ \cline{2-6}
% & \textbf{HI} & 48.31 & 63.32 & 48.31 & 43.73 \\ \cline{2-6}
% & \textbf{UR} & 47.06 & 61.61 & 47.06 & 42.62 \\ \midrule

% \multirow{4}{*}{\textbf{Gemini}} & \textbf{EN} & 64.59 & 67.86 & 64.59 & 64.44 \\ \cline{2-6}
% & \textbf{BN} & 65.40 & 66.68 & 65.40 & \textbf{64.93} \\ \cline{2-6}
% & \textbf{HI} & 65.87 & 67.14 & 65.87 & \textbf{65.33} \\ \cline{2-6}
% & \textbf{UR} & 65.93 & 66.77 & 65.93 & \textbf{65.14} \\

% \bottomrule
% \end{tabular}
% }
% \caption{Performances of the Sentiment task across the models and languages. \textbf{Bold} indicates the best performances across the languages. Lang.: language, Acc.: accuracy, P.: Precision, R.: Recall, EN: English, BN: Bangla, HI: Hindi, and UR: Urdu}
% \label{tab:performance_senti}
% \end{table}

We postulate the low performance of LLMs in low-resource languages for the following reasons. One of the main reasons is that most of the LLMs are trained on a large amount ($90\%$) of English data, %of the training data of Llama 2 is English, 
whereas the amount of training data for low-resource languages is small compared with English. Moreover, cultural differences between English-spoken countries and low-resource language countries affect the sentiment and hate speech tasks the most. Lastly, the quality of the translation affects the performance of low-resource languages. However, Hindi performed better than Bangla and Urdu in all tasks among the low-resource languages. The performance difference among the low-resource languages is insignificant across the tasks and LLMs. Our findings from this section conclude that improving LLMs is required for low-resource languages.

% \subsection{Comparison Among LLMs}
%How do LLMs perform?

\begin{figure}[h]
     \centering
     \includegraphics[width=0.45\textwidth ]{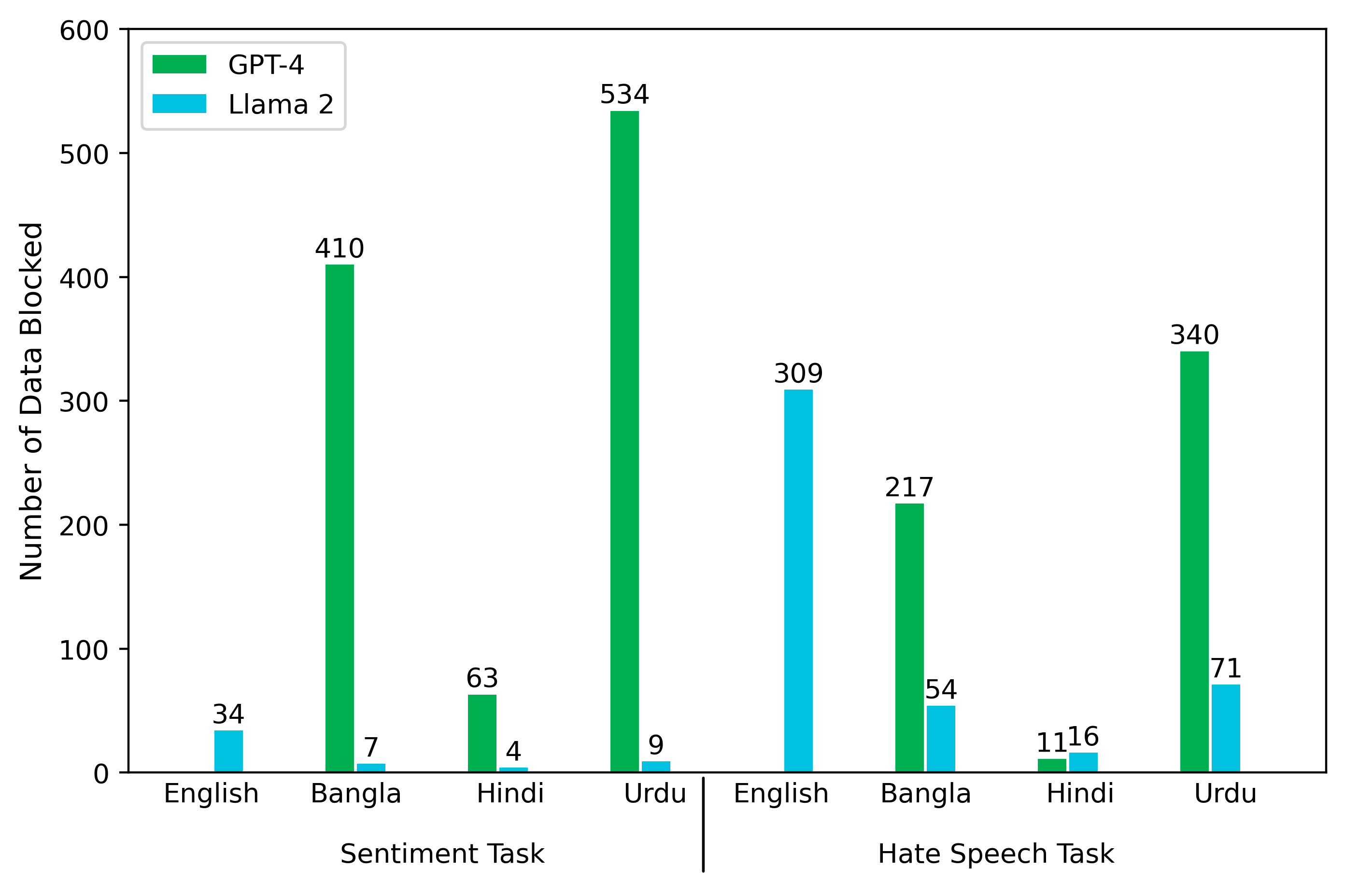}
     \caption{Number of unpredicted samples by GPT-4 and Llama 2. Note that we only include the languages and models from the tasks with unpredicted samples.}
     \label{fig:unpredicted}
 \end{figure}

\noindent\textbf{Comparison Among LLMs:} We first analyzed the individual LLM outputs and found that GPT-4 could not predict much data on sentiment and hate speech tasks for Bangla and Urdu. Moreover, GPT-4 was able to provide predictions for all the English language samples for all the tasks. We also noticed that Llama 2 and Gemini models could predict all the samples from the NLI task for all languages. Llama 2 could not predict much data on the hate speech task for English. However, Llama 2 provides a small number of unpredicted data compared with GPT-4 for Bangla, Hindi, and Urdu. We analyzed the response of unpredicted data from GPT-4. We found that the model cannot understand the context to classify while Llama 2 could not predict due to inappropriate or offensive language. Moreover, some responses of Llama include repeated `l' as the label.
We briefly overview the unpredicted data in  Figure \ref{fig:unpredicted}. During the evaluation metrics calculation, we assigned the inverse classes for the unpredicted samples.

% \begin{table}[!ht]
% \centering
% \resizebox{\linewidth}{!}{
% \begin{tabular}{llrrrr}
% \toprule
% \multicolumn{1}{c}{\textbf{Model}} & \multicolumn{1}{c}{\textbf{Lang.}} & \multicolumn{1}{c}{\textbf{Acc.}} & \multicolumn{1}{c}{\textbf{P.}} & \multicolumn{1}{c}{\textbf{R.}} & \multicolumn{1}{c}{\textbf{F1$_{macro}$}}\\ \midrule

% \multirow{4}{*}{\textbf{GPT-4}} & \textbf{EN} & 86.81 & 85.52 & 86.81 & \textbf{62.54} \\ \cline{2-6}
% & \textbf{BN} & 55.32 & 75.51 & 55.32 & 38.79 \\ \cline{2-6}
% & \textbf{HI} & 64.66 & 77.93 & 64.66 & \textbf{44.61} \\ \cline{2-6}
% & \textbf{UR} & 54.00 & 75.18 & 54.00 & 38.66 \\ \midrule

% \multirow{4}{*}{\textbf{Llama 2}} & \textbf{EN} & 79.32 & 83.93 & 79.32 & 60.04 \\ \cline{2-6}
% & \textbf{BN} & 69.92 & 69.12 & 69.92 & \textbf{41.36} \\ \cline{2-6}
% & \textbf{HI} & 74.54 & 71.58 & 74.54 & 44.39 \\ \cline{2-6}
% & \textbf{UR} & 47.29 & 65.68 & 47.29 & \textbf{43.54} \\ \midrule

% \multirow{4}{*}{\textbf{Gemini}} & \textbf{EN} & 58.00 & 77.69 & 58.00 & 49.10 \\ \cline{2-6}
% & \textbf{BN} & 30.34 & 70.93 & 30.34 & 30.81 \\ \cline{2-6}
% & \textbf{HI} & 32.01 & 72.72 & 32.01 & 33.36 \\ \cline{2-6}
% & \textbf{UR} & 28.56 & 70.07 & 28.56 & 28.47 \\

% \bottomrule
% \end{tabular}
% }
% \caption{Performances of the Hate speech task across the models and languages. \textbf{Bold} indicates the best performances across the languages. Lang.: language, Acc.: accuracy, P.: Precision, R.: Recall, EN: English, BN: Bangla, HI: Hindi, and UR: Urdu}
% \label{tab:performance_hate}
% \end{table}

\begin{figure}[h]
     \centering
     \includegraphics[width=0.45\textwidth ]{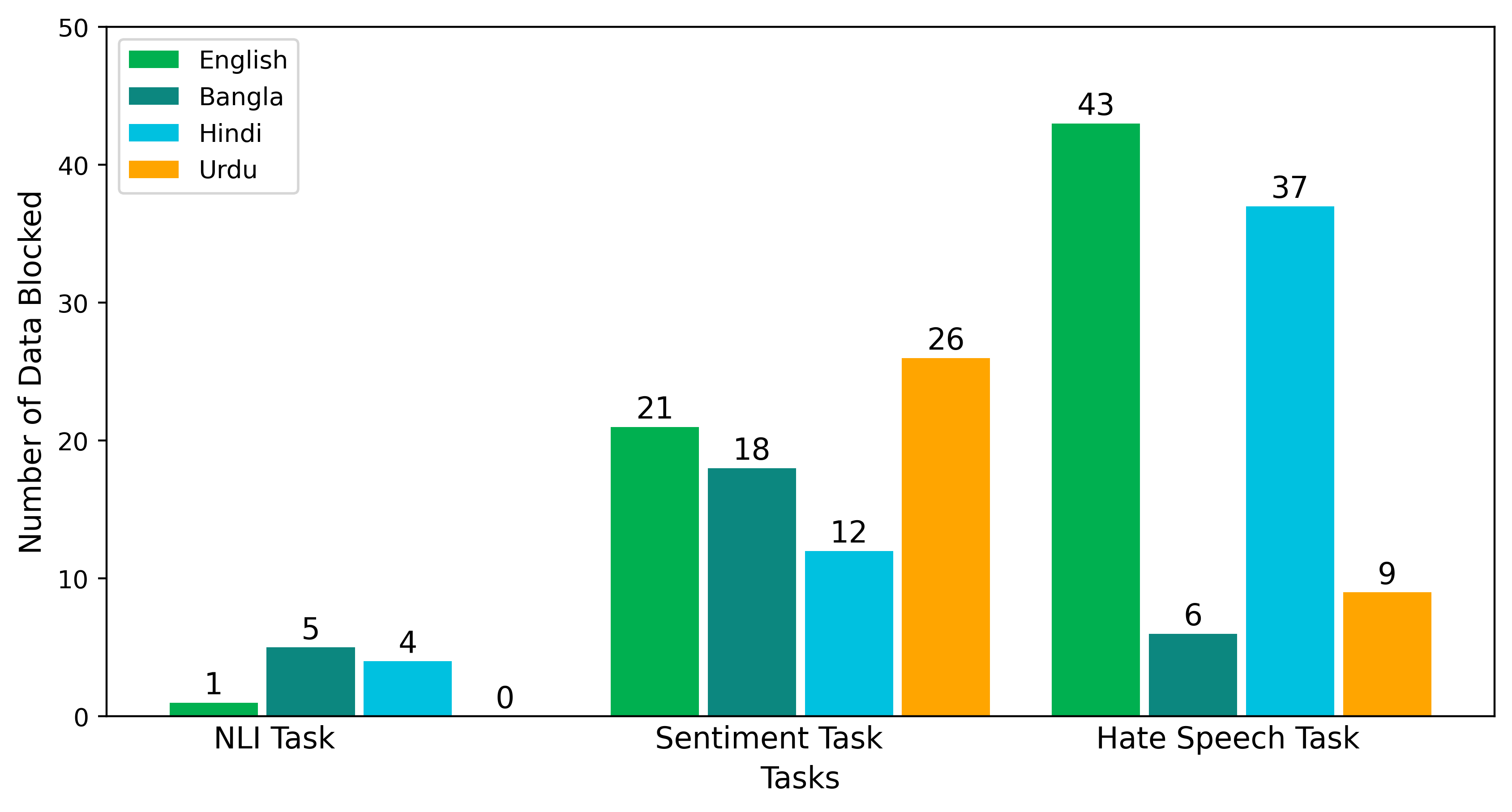}
     \caption{Number of samples that are blocked by Gemini.}
     \label{fig:blocked}
 \end{figure}

Gemini is the only LLM that predicted all the samples of each task. Although we provide a safety setting for the Gemini model, it blocked some data due to the content containing derogatory language. We noticed that the samples from sentiment and hate speech tasks were blocked for containing derogatory language, and those from the NLI task were not blocked. We provide a brief overview of the number of samples that are blocked by Gemini in Figure \ref{fig:blocked}. However, the Urdu language is not supported by the Gemini. Despite that, the Gemini performs strongly in Urdu for the NLI and sentiment tasks. We further investigated the performances of Gemini in the Urdu language. We found that the alphabets of Urdu are derived from the Arabic language family\footnote{\url{https://en.wikipedia.org/wiki/Urdu\_alphabet}} and many words are adopted from the Arabic language. Arabic is supported by Gemini, and the training data of Arabic shares semantic information with the Urdu language, which is why Gemini exhibits a strong performance in the Urdu language.

%Further, we investigated the detailed performances of each task. GPT-4 shows superior performances on the NLI task for all languages while exhibiting good performances on the sentiment task. However, most hate class data were misclassified in the hate speech task for all languages. Llama 2 provides strong performances in English for NLI, sentiment, and hate speech tasks while finding difficulties in accurately predicting the contradiction, neutral, and hate classes for NLI, sentiment, and hate speech tasks, respectively. Although Llama 2 outperforms GPT-4 performances in hate class in every language, GPT-4 in English and Hindi is better than Llama 2 for hate speech tasks. Moreover, Llama 2 demonstrated comparatively better performance on the hate speech task than NLI and sentiment tasks. While Gemini exhibits strong performances in NLI and sentiment tasks for all the languages, it consistently performs poorly on the speech task for all the languages. However, Gemini performs comparatively better hate class performance than Llama 2 and GPT-4 for all the languages. Moreover, the performances in the neither and offensive classes are worse than other LLMs. We also found that most offensive classes are misclassified as neither. We provided the detailed class-wise experimental results in Appendix \ref{sec:appendix_b}.

In general, GPT-4 shows prominent performances over other LLMs across all the tasks. Although Llama 2 provides better results for hate speech tasks, it struggled to perform well in NLI and sentiment tasks. While Gemini demonstrated strong performances in NLI and sentiment tasks, it delivered worse in hate speech tasks. Despite observing a smaller performance gap in Gemini, significant disparities persist in GPT-4 and Llama-2, indicating that direct translation is less likely to compromise sentiment information. See Appendix \ref{sec:appendix_b} for class-wise experimental results.

\noindent\textbf{Tasks Performances:} The overall performance of the NLI task is comparatively better than sentiment and hate speech tasks (Table~\ref{tab:performance_all}). The definition of an NLI task has clear rules and structured patterns, while sentiment and hate speech tasks are subjective and context-dependent. NLI task identifies the relation between two sentences based on structure and language logic \cite{bowman2015large} that makes the task easier for LLMs. Moreover, the context lies with the sentence pair, and LLMs can understand the context. While sentiment and hate speech tasks require understanding the tone of the text and sometimes the complex social and cultural contexts, these facts are challenging for LLMs to understand. 
Moreover, the data of the NLI task is incorporated from the well-structured MNLI corpus with precise labels and balanced classes, making the task more comfortable for LLMs. Unlike the NLI task, sentiment and hate speech task data are curated from social media platforms containing noise, informal expressions, slang, and incomplete text, making it challenging for LLMs. Moreover, most of the texts do not have the contexts within their representation, and it is challenging to identify the context for both humans and LLMs. Straightforward linguistics features and contextual information make the NLI task easier and perform better than sentiment and hate speech tasks using different LLMs. In addition, during the evaluation, we explored whether English hashtags have any impact on predictions for Bangla, Hindi, and Urdu. Our empirical results demonstrated that LLMs do not rely solely on hashtags but on the entire sequence.

\section{Conclusion}
\label{sec:conclusion}

In this study, we introduce datasets for sentiment and hate speech tasks by translating from English to Bangla, Hindi, and Urdu to facilitate research in low-resource language processing. Through a comprehensive examination of zero-shot learning across multiple LLMs, notably GPT-4, we uncover performance disparities between English and low-resource languages. Furthermore, our analysis identifies NLI as a task where GPT-4 consistently demonstrates superior capabilities, underscoring avenues for enhancing LLM applicability in general-purpose NLP applications.

% In conclusion, our comprehensive analysis sheds light on the pivotal role of large language models (LLMs) in the landscape of natural language processing (NLP), emphasizing both their remarkable performance in resource-rich languages such as English and the pressing need to extend their utility to low-resource language settings. Through our investigation of zero-shot learning with various LLMs, we have demonstrated that while LLMs, notably GPT-4, exhibit commendable capabilities in English, their performance in low-resource languages remains a subject of concern. 
% %This study underscores the importance of addressing the dearth of research and evaluation in low-resource language contexts, propelled by constraints including dataset scarcity and computational expenses. 
% Our findings not only highlight the existing gap in linguistic capabilities for low-resource languages but also advocate for concerted efforts to bridge this divide. By focusing on tasks such as natural language inference (NLI) and considering performance across different LLM architectures, our research contributes valuable insights into the potential avenues for enhancing the applicability of LLMs in general-purpose NLP applications. Moving forward, concerted interdisciplinary efforts are warranted to bolster research initiatives aimed at refining LLM performance in low-resource language environments, thus fostering inclusivity and accessibility in the realm of natural language processing.

\clearpage
\section*{Limitation}
%\textcolor{red}{This section is now mandatory from February ARR.}

In our study, we refrained from utilizing explicit prompting techniques to enhance the performance of large language models (LLMs). Our evaluation primarily focused on assessing LLMs in the context of English and low-resource languages such as Bangla, Hindi, and Urdu, without exploring variations in prompts. Regarding the quality of dataset translations, it is important to note that the translations generated by Google Translator were not subjected to human verification. Consequently, while certain translation errors were overlooked during our analysis, we conducted sampling from each translated dataset to gain insights into the overall translation quality. Our findings underscore the necessity for further refinement in translation methodologies to elevate both the quality and accuracy of translations in future research endeavors.

% \paragraph{Prompting} We did not use explicit prompting techniques to achieve better performances of LLMs. Our study mainly evaluates LLMs for English and low-resource languages such as Bangla, Hindi, and Urdu. Also, different prompts were not explored during our study.

% \paragraph{Dataset Translation Quality} The translations generated by Google Translator were not verified by humans. As a result, we ignore some translation errors during the study. 
% However, we sampled a few data from each translated dataset to understand translation quality in summary.  Our findings suggests the need for further refinement in translation methodologies to enhance overall quality and accuracy in future works.

\bibliography{custom}

\begin{thebibliography}{46}
\expandafter\ifx\csname natexlab\endcsname\relax\def\natexlab#1{#1}\fi

\bibitem[{Aggarwal et~al.(2022)Aggarwal, Gupta, and Kunchukuttan}]{aggarwal2022indicxnli}
Divyanshu Aggarwal, Vivek Gupta, and Anoop Kunchukuttan. 2022.
\newblock Indicxnli: Evaluating multilingual inference for indian languages.
\newblock \emph{arXiv preprint arXiv:2204.08776}.

\bibitem[{Ahuja et~al.(2023{\natexlab{a}})Ahuja, Diddee, Hada, Ochieng, Ramesh, Jain, Nambi, Ganu, Segal, Axmed et~al.}]{ahuja2023mega}
Kabir Ahuja, Harshita Diddee, Rishav Hada, Millicent Ochieng, Krithika Ramesh, Prachi Jain, Akshay Nambi, Tanuja Ganu, Sameer Segal, Maxamed Axmed, et~al. 2023{\natexlab{a}}.
\newblock Mega: Multilingual evaluation of generative ai.
\newblock \emph{arXiv preprint arXiv:2303.12528}.

\bibitem[{Ahuja et~al.(2023{\natexlab{b}})Ahuja, Aggarwal, Gumma, Watts, Sathe, Ochieng, Hada, Jain, Axmed, Bali et~al.}]{ahuja2023megaverse}
Sanchit Ahuja, Divyanshu Aggarwal, Varun Gumma, Ishaan Watts, Ashutosh Sathe, Millicent Ochieng, Rishav Hada, Prachi Jain, Maxamed Axmed, Kalika Bali, et~al. 2023{\natexlab{b}}.
\newblock Megaverse: benchmarking large language models across languages, modalities, models and tasks.
\newblock \emph{arXiv preprint arXiv:2311.07463}.

\bibitem[{Akter et~al.(2023)Akter, Yu, Muhamed, Ou, B{\"a}uerle, Cabrera, Dholakia, Xiong, and Neubig}]{akter2023depth}
Syeda~Nahida Akter, Zichun Yu, Aashiq Muhamed, Tianyue Ou, Alex B{\"a}uerle, {\'A}ngel~Alexander Cabrera, Krish Dholakia, Chenyan Xiong, and Graham Neubig. 2023.
\newblock An in-depth look at gemini's language abilities.
\newblock \emph{arXiv preprint arXiv:2312.11444}.

\bibitem[{Alam et~al.(2021)Alam, Hasan, Alam, Khan, Tajrin, Khan, and Chowdhury}]{alam2021review}
Firoj Alam, Arid Hasan, Tanvirul Alam, Akib Khan, Janntatul Tajrin, Naira Khan, and Shammur~Absar Chowdhury. 2021.
\newblock A review of bangla natural language processing tasks and the utility of transformer models.
\newblock \emph{arXiv preprint arXiv:2107.03844}.

\bibitem[{Asai et~al.(2023)Asai, Kudugunta, Yu, Blevins, Gonen, Reid, Tsvetkov, Ruder, and Hajishirzi}]{asai2023buffet}
Akari Asai, Sneha Kudugunta, Xinyan~Velocity Yu, Terra Blevins, Hila Gonen, Machel Reid, Yulia Tsvetkov, Sebastian Ruder, and Hannaneh Hajishirzi. 2023.
\newblock Buffet: Benchmarking large language models for few-shot cross-lingual transfer.
\newblock \emph{arXiv preprint arXiv:2305.14857}.

\bibitem[{Bhattacharjee et~al.(2021)Bhattacharjee, Hasan, Samin, Islam, Rahman, Iqbal, and Shahriyar}]{bhattacharjee2021banglabert}
Abhik Bhattacharjee, Tahmid Hasan, Kazi Samin, Md~Saiful Islam, M.~Sohel Rahman, Anindya Iqbal, and Rifat Shahriyar. 2021.
\newblock \href {http://arxiv.org/abs/2101.00204} {Banglabert: Combating embedding barrier in multilingual models for low-resource language understanding}.

\bibitem[{Bowman et~al.(2015)Bowman, Angeli, Potts, and Manning}]{bowman2015large}
Samuel~R Bowman, Gabor Angeli, Christopher Potts, and Christopher~D Manning. 2015.
\newblock A large annotated corpus for learning natural language inference.
\newblock \emph{arXiv preprint arXiv:1508.05326}.

\bibitem[{Conneau et~al.(2018)Conneau, Rinott, Lample, Williams, Bowman, Schwenk, and Stoyanov}]{conneau2018xnli}
Alexis Conneau, Ruty Rinott, Guillaume Lample, Adina Williams, Samuel~R. Bowman, Holger Schwenk, and Veselin Stoyanov. 2018.
\newblock Xnli: Evaluating cross-lingual sentence representations.
\newblock In \emph{Proceedings of the 2018 Conference on Empirical Methods in Natural Language Processing}. Association for Computational Linguistics.

\bibitem[{Davidson et~al.(2017)Davidson, Warmsley, Macy, and Weber}]{hateoffensive}
Thomas Davidson, Dana Warmsley, Michael Macy, and Ingmar Weber. 2017.
\newblock Automated hate speech detection and the problem of offensive language.
\newblock In \emph{Proceedings of the 11th International AAAI Conference on Web and Social Media}, ICWSM '17, pages 512--515.

\bibitem[{Garc{\'\i}a-D{\'\i}az et~al.(2023)Garc{\'\i}a-D{\'\i}az, Pan, and Valencia-Garc{\'\i}a}]{garcia2023leveraging}
Jos{\'e}~Antonio Garc{\'\i}a-D{\'\i}az, Ronghao Pan, and Rafael Valencia-Garc{\'\i}a. 2023.
\newblock Leveraging zero and few-shot learning for enhanced model generality in hate speech detection in spanish and english.
\newblock \emph{Mathematics}, 11(24):5004.

\bibitem[{Gubelmann et~al.(2023)Gubelmann, Kalouli, Niklaus, and Handschuh}]{gubelmann2023truth}
Reto Gubelmann, Aikaterini-Lida Kalouli, Christina Niklaus, and Siegfried Handschuh. 2023.
\newblock When truth matters-addressing pragmatic categories in natural language inference (nli) by large language models (llms).
\newblock In \emph{Proceedings of the 12th Joint Conference on Lexical and Computational Semantics (* SEM 2023)}, pages 24--39.

\bibitem[{Hasan et~al.(2023{\natexlab{a}})Hasan, Alam, Anjum, Das, and Anjum}]{hasan2023blp}
Md~Arid Hasan, Firoj Alam, Anika Anjum, Shudipta Das, and Afiyat Anjum. 2023{\natexlab{a}}.
\newblock Blp-2023 task 2: Sentiment analysis.
\newblock In \emph{Proceedings of the First Workshop on Bangla Language Processing (BLP-2023)}, pages 354--364.

\bibitem[{Hasan et~al.(2023{\natexlab{b}})Hasan, Das, Anjum, Alam, Anjum, Sarker, and Noori}]{hasan2023zero}
Md~Arid Hasan, Shudipta Das, Afiyat Anjum, Firoj Alam, Anika Anjum, Avijit Sarker, and Sheak Rashed~Haider Noori. 2023{\natexlab{b}}.
\newblock Zero-and few-shot prompting with llms: A comparative study with fine-tuned models for bangla sentiment analysis.
\newblock \emph{arXiv preprint arXiv:2308.10783}.

\bibitem[{Hasan et~al.(2020)Hasan, Tajrin, Chowdhury, and Alam}]{hasan2020sentiment}
Md~Arid Hasan, Jannatul Tajrin, Shammur~Absar Chowdhury, and Firoj Alam. 2020.
\newblock Sentiment classification in bangla textual content: A comparative study.
\newblock In \emph{2020 23rd international conference on computer and information technology (ICCIT)}, pages 1--6. IEEE.

\bibitem[{Hee et~al.(2024)Hee, Sharma, Cao, Nandi, Nakov, Chakraborty, and Lee}]{hee2024recent}
Ming~Shan Hee, Shivam Sharma, Rui Cao, Palash Nandi, Preslav Nakov, Tanmoy Chakraborty, and Roy Ka-Wei Lee. 2024.
\newblock Recent advances in hate speech moderation: Multimodality and the role of large models.
\newblock \emph{arXiv preprint arXiv:2401.16727}.

\bibitem[{Holmstr{\"o}m et~al.(2023)Holmstr{\"o}m, Kunz, and Kuhlmann}]{holmstrom2023bridging}
Oskar Holmstr{\"o}m, Jenny Kunz, and Marco Kuhlmann. 2023.
\newblock Bridging the resource gap: Exploring the efficacy of english and multilingual llms for swedish.
\newblock In \emph{Proceedings of the Second Workshop on Resources and Representations for Under-Resourced Languages and Domains (RESOURCEFUL-2023)}, pages 92--110.

\bibitem[{Hu et~al.(2020)Hu, Ruder, Siddhant, Neubig, Firat, and Johnson}]{hu2020xtreme}
Junjie Hu, Sebastian Ruder, Aditya Siddhant, Graham Neubig, Orhan Firat, and Melvin Johnson. 2020.
\newblock Xtreme: A massively multilingual multi-task benchmark for evaluating cross-lingual generalisation.
\newblock In \emph{International Conference on Machine Learning}, pages 4411--4421. PMLR.

\bibitem[{Huang et~al.(2023)Huang, Tang, Zhang, Zhao, Song, Xia, and Wei}]{huang2023not}
Haoyang Huang, Tianyi Tang, Dongdong Zhang, Wayne~Xin Zhao, Ting Song, Yan Xia, and Furu Wei. 2023.
\newblock Not all languages are created equal in llms: Improving multilingual capability by cross-lingual-thought prompting.
\newblock \emph{arXiv preprint arXiv:2305.07004}.

\bibitem[{Islam et~al.(2021)Islam, Kar, Islam, and Amin}]{islam2021sentnob}
Khondoker~Ittehadul Islam, Sudipta Kar, Md~Saiful Islam, and Mohammad~Ruhul Amin. 2021.
\newblock Sentnob: A dataset for analysing sentiment on noisy bangla texts.
\newblock In \emph{Findings of the Association for Computational Linguistics: EMNLP 2021}, pages 3265--3271.

\bibitem[{Kabir et~al.(2023)Kabir, Islam, Laskar, Nayeem, Bari, and Hoque}]{kabir2023benllmeval}
Mohsinul Kabir, Mohammed~Saidul Islam, Md~Tahmid~Rahman Laskar, Mir~Tafseer Nayeem, M~Saiful Bari, and Enamul Hoque. 2023.
\newblock Benllmeval: A comprehensive evaluation into the potentials and pitfalls of large language models on bengali nlp.
\newblock \emph{arXiv preprint arXiv:2309.13173}.

\bibitem[{Khan et~al.(2021)Khan, Shahzad, and Malik}]{khan2021hate}
Muhammad~Moin Khan, Khurram Shahzad, and Muhammad~Kamran Malik. 2021.
\newblock Hate speech detection in roman urdu.
\newblock \emph{ACM Transactions on Asian and Low-Resource Language Information Processing (TALLIP)}, 20(1):1--19.

\bibitem[{Koto et~al.(2024)Koto, Beck, Talat, Gurevych, and Baldwin}]{koto2024zero}
Fajri Koto, Tilman Beck, Zeerak Talat, Iryna Gurevych, and Timothy Baldwin. 2024.
\newblock Zero-shot sentiment analysis in low-resource languages using a multilingual sentiment lexicon.
\newblock \emph{arXiv preprint arXiv:2402.02113}.

\bibitem[{Kumar and Albuquerque(2021)}]{kumar2021sentiment}
Akshi Kumar and Victor Hugo~C Albuquerque. 2021.
\newblock Sentiment analysis using xlm-r transformer and zero-shot transfer learning on resource-poor indian language.
\newblock \emph{Transactions on Asian and Low-Resource Language Information Processing}, 20(5):1--13.

\bibitem[{Lai et~al.(2023)Lai, Van~Nguyen, Ngo, Nguyen, Dernoncourt, Rossi, and Nguyen}]{lai2023okapi}
Viet~Dac Lai, Chien Van~Nguyen, Nghia~Trung Ngo, Thuat Nguyen, Franck Dernoncourt, Ryan~A Rossi, and Thien~Huu Nguyen. 2023.
\newblock Okapi: Instruction-tuned large language models in multiple languages with reinforcement learning from human feedback.
\newblock \emph{arXiv preprint arXiv:2307.16039}.

\bibitem[{Lu et~al.(2023)Lu, Bansal, Xia, Liu, Li, Hajishirzi, Cheng, Chang, Galley, and Gao}]{lu2023mathvista}
Pan Lu, Hritik Bansal, Tony Xia, Jiacheng Liu, Chunyuan Li, Hannaneh Hajishirzi, Hao Cheng, Kai-Wei Chang, Michel Galley, and Jianfeng Gao. 2023.
\newblock Mathvista: Evaluating mathematical reasoning of foundation models in visual contexts.
\newblock \emph{arXiv preprint arXiv:2310.02255}.

\bibitem[{Lyu et~al.(2023)Lyu, Xu, and Wang}]{lyu2023new}
Chenyang Lyu, Jitao Xu, and Longyue Wang. 2023.
\newblock New trends in machine translation using large language models: Case examples with chatgpt.
\newblock \emph{arXiv preprint arXiv:2305.01181}.

\bibitem[{Miao et~al.(2023)Miao, Teh, and Rainforth}]{miao2023selfcheck}
Ning Miao, Yee~Whye Teh, and Tom Rainforth. 2023.
\newblock Selfcheck: Using llms to zero-shot check their own step-by-step reasoning.
\newblock \emph{arXiv preprint arXiv:2308.00436}.

\bibitem[{OpenAI(2023)}]{openai2023gpt}
R~OpenAI. 2023.
\newblock Gpt-4 technical report.
\newblock \emph{arXiv}, pages 2303--08774.

\bibitem[{Pahwa and Pahwa(2023)}]{pahwa2023bphigh}
Bhavish Pahwa and Bhavika Pahwa. 2023.
\newblock Bphigh at semeval-2023 task 7: Can fine-tuned cross-encoders outperform gpt-3.5 in nli tasks on clinical trial data?
\newblock In \emph{Proceedings of the 17th International Workshop on Semantic Evaluation (SemEval-2023)}, pages 1936--1944.

\bibitem[{Qin et~al.(2023)Qin, Chen, Wei, Huang, and Che}]{qin2023cross}
Libo Qin, Qiguang Chen, Fuxuan Wei, Shijue Huang, and Wanxiang Che. 2023.
\newblock Cross-lingual prompting: Improving zero-shot chain-of-thought reasoning across languages.
\newblock \emph{arXiv preprint arXiv:2310.14799}.

\bibitem[{Rane(2023)}]{rane2023enhancing}
Nitin Rane. 2023.
\newblock Enhancing mathematical capabilities through chatgpt and similar generative artificial intelligence: Roles and challenges in solving mathematical problems.
\newblock \emph{Available at SSRN 4603237}.

\bibitem[{Robinson et~al.(2023)Robinson, Ogayo, Mortensen, and Neubig}]{robinson2023chatgpt}
Nathaniel~R Robinson, Perez Ogayo, David~R Mortensen, and Graham Neubig. 2023.
\newblock Chatgpt mt: Competitive for high-(but not low-) resource languages.
\newblock \emph{arXiv preprint arXiv:2309.07423}.

\bibitem[{Rosenthal et~al.(2017)Rosenthal, Farra, and Nakov}]{SemEval:2017:task4}
Sara Rosenthal, Noura Farra, and Preslav Nakov. 2017.
\newblock {SemEval}-2017 task 4: Sentiment analysis in {T}witter.
\newblock In \emph{Proceedings of the 11th International Workshop on Semantic Evaluation}, SemEval '17, Vancouver, Canada. Association for Computational Linguistics.

\bibitem[{Ruder et~al.(2021)Ruder, Constant, Botha, Siddhant, Firat, Fu, Liu, Hu, Garrette, Neubig et~al.}]{ruder2021xtreme}
Sebastian Ruder, Noah Constant, Jan Botha, Aditya Siddhant, Orhan Firat, Jinlan Fu, Pengfei Liu, Junjie Hu, Dan Garrette, Graham Neubig, et~al. 2021.
\newblock Xtreme-r: Towards more challenging and nuanced multilingual evaluation.
\newblock \emph{arXiv preprint arXiv:2104.07412}.

\bibitem[{Santosh and Aravind(2019)}]{santosh2019hate}
TYSS Santosh and KVS Aravind. 2019.
\newblock Hate speech detection in hindi-english code-mixed social media text.
\newblock In \emph{Proceedings of the ACM India joint international conference on data science and management of data}, pages 310--313.

\bibitem[{Sun et~al.(2023)Sun, Li, Zhang, Wang, Wu, Li, Zhang, and Wang}]{sun2023sentiment}
Xiaofei Sun, Xiaoya Li, Shengyu Zhang, Shuhe Wang, Fei Wu, Jiwei Li, Tianwei Zhang, and Guoyin Wang. 2023.
\newblock Sentiment analysis through llm negotiations.
\newblock \emph{arXiv preprint arXiv:2311.01876}.

\bibitem[{Suzgun et~al.(2022)Suzgun, Scales, Sch{\"a}rli, Gehrmann, Tay, Chung, Chowdhery, Le, Chi, Zhou et~al.}]{suzgun2022challenging}
Mirac Suzgun, Nathan Scales, Nathanael Sch{\"a}rli, Sebastian Gehrmann, Yi~Tay, Hyung~Won Chung, Aakanksha Chowdhery, Quoc~V Le, Ed~H Chi, Denny Zhou, et~al. 2022.
\newblock Challenging big-bench tasks and whether chain-of-thought can solve them.
\newblock \emph{arXiv preprint arXiv:2210.09261}.

\bibitem[{Tan et~al.(2023)Tan, Min, Li, Li, Hu, Chen, and Qi}]{tan2023can}
Yiming Tan, Dehai Min, Yu~Li, Wenbo Li, Nan Hu, Yongrui Chen, and Guilin Qi. 2023.
\newblock Can chatgpt replace traditional kbqa models? an in-depth analysis of the question answering performance of the gpt llm family.
\newblock In \emph{International Semantic Web Conference}, pages 348--367. Springer.

\bibitem[{Team et~al.(2023)Team, Anil, Borgeaud, Wu, Alayrac, Yu, Soricut, Schalkwyk, Dai, Hauth et~al.}]{team2023gemini}
Gemini Team, Rohan Anil, Sebastian Borgeaud, Yonghui Wu, Jean-Baptiste Alayrac, Jiahui Yu, Radu Soricut, Johan Schalkwyk, Andrew~M Dai, Anja Hauth, et~al. 2023.
\newblock Gemini: a family of highly capable multimodal models.
\newblock \emph{arXiv preprint arXiv:2312.11805}.

\bibitem[{Touvron et~al.(2023)Touvron, Martin, Stone, Albert, Almahairi, Babaei, Bashlykov, Batra, Bhargava, Bhosale et~al.}]{touvron2023llama}
Hugo Touvron, Louis Martin, Kevin Stone, Peter Albert, Amjad Almahairi, Yasmine Babaei, Nikolay Bashlykov, Soumya Batra, Prajjwal Bhargava, Shruti Bhosale, et~al. 2023.
\newblock Llama 2: Open foundation and fine-tuned chat models.
\newblock \emph{arXiv preprint arXiv:2307.09288}.

\bibitem[{Xing(2024)}]{xing2024designing}
Frank Xing. 2024.
\newblock Designing heterogeneous llm agents for financial sentiment analysis.
\newblock \emph{arXiv preprint arXiv:2401.05799}.

\bibitem[{Xu et~al.(2023)Xu, Kim, Sharaf, and Awadalla}]{xu2023paradigm}
Haoran Xu, Young~Jin Kim, Amr Sharaf, and Hany~Hassan Awadalla. 2023.
\newblock A paradigm shift in machine translation: Boosting translation performance of large language models.
\newblock \emph{arXiv preprint arXiv:2309.11674}.

\bibitem[{Zhang et~al.(2023{\natexlab{a}})Zhang, Yang, and Liu}]{zhang2023instruct}
Boyu Zhang, Hongyang Yang, and Xiao-Yang Liu. 2023{\natexlab{a}}.
\newblock Instruct-fingpt: Financial sentiment analysis by instruction tuning of general-purpose large language models.
\newblock \emph{arXiv preprint arXiv:2306.12659}.

\bibitem[{Zhang et~al.(2023{\natexlab{b}})Zhang, Yang, Zhou, Ali~Babar, and Liu}]{zhang2023enhancing}
Boyu Zhang, Hongyang Yang, Tianyu Zhou, Muhammad Ali~Babar, and Xiao-Yang Liu. 2023{\natexlab{b}}.
\newblock Enhancing financial sentiment analysis via retrieval augmented large language models.
\newblock In \emph{Proceedings of the Fourth ACM International Conference on AI in Finance}, pages 349--356.

\bibitem[{Zhuang et~al.(2023)Zhuang, Yu, Wang, Sun, and Zhang}]{zhuang2023toolqa}
Yuchen Zhuang, Yue Yu, Kuan Wang, Haotian Sun, and Chao Zhang. 2023.
\newblock Toolqa: A dataset for llm question answering with external tools.
\newblock \emph{arXiv preprint arXiv:2306.13304}.

\end{thebibliography}
\bibliographystyle{acl_natbib}

\appendix
\section{Prompts and Safety Setting}
\label{sec:appendix_a}

This section presents the details of the prompts that we used for each model and task\footnote{Note that we use the same prompt for each task.}. We present the example prompt for the NLI task, sentiment task, and Hatespeech task in Table \ref{tab:prompt1}, Table \ref{tab:prompt2}, and Table \ref{tab:prompt3} respectively. We provide the details of the safety setting for the Gemini Pro model in Table \ref{tab:safety_setting}

\begin{table}[!ht]
\centering
\resizebox{\linewidth}{!}{
\begin{tabular}{p{1.9cm}|p{7cm}}
\hline
\textbf{Model} & \textbf{Prompt} \\ \hline

\textbf{GPT-4} & 
[
\{ 

`role': `user',

`content': "Classify the following `premise' and `hypothesis' into one of the following classes: `Entailment', `Contradiction', or `Neutral'. Provide only label as your response."

premise: [PREMISE\_TEXT]

hypothesis: [HYPOTHESIS\_TEXT]

label:

\},

\{

role: `system',

content: "You are an expert data annotator and your task is to analyze the text and find the appropriate output that is defined in the user content."

\}
]
\\ \hline
\textbf{Llama 2 and Gemini} & 
Classify the following `premise' and `hypothesis' into one of the following classes: `Entailment', `Contradiction', or `Neutral'. Provide only label as your response.

premise: [PREMISE\_TEXT]

hypothesis: [HYPOTHESIS\_TEXT]

label:
\\
\bottomrule
\end{tabular}
}
\caption{Prompts used for zero-shot learning in NLI task.}
\label{tab:prompt1}
\end{table}

\begin{table}[!ht]
\centering
\resizebox{\linewidth}{!}{
\begin{tabular}{p{1.9cm}|p{7cm}}
\hline
\textbf{Model} & \textbf{Prompt} \\ \hline

\textbf{GPT-4} & 
[
\{ 

`role': `user',

`content': "Classify the `text' into one of the following labels: `Positive', `Neutral', or `Negative'. Provide only label as your response."

text: [SOURCE\_TEXT]

label:

\},

\{

role: `system',

content: "You are an expert data annotator and your task is to analyze the text and find the appropriate output that is defined in the user content."

\}
]
\\ \hline
\textbf{Llama 2 and Gemini} & 
Classify the `text' into one of the following labels: `Positive', `Neutral', or `Negative'. Provide only label as your response.

text: [SOURCE\_TEXT]

label:
\\
\bottomrule
\end{tabular}
}
\caption{Prompts used for zero-shot learning in Sentiment task.}
\label{tab:prompt2}
\end{table}

\begin{table}[!ht]
\centering
\resizebox{\linewidth}{!}{
\begin{tabular}{p{1.9cm}|p{7cm}}
\hline
\textbf{Model} & \textbf{Prompt} \\ \hline

\textbf{GPT-4} & 
[
\{ 

`role': `user',

`content': "Classify the `text' into one of the following labels: `Hate', `Offensive', or `Neither'. Provide only label as your response."

text: [SOURCE\_TEXT]

label:

\},

\{

role: `system',

content: "You are an expert data annotator and your task is to analyze the text and find the appropriate output that is defined in the user content."

\}
]
\\ \hline
\textbf{Llama 2 and Gemini} & 
Classify the `text' into one of the following labels: `Hate', `Offensive', or `Neither'. Provide only label as your response.

text: [SOURCE\_TEXT]

label:
\\
\bottomrule
\end{tabular}
}
\caption{Prompts used for zero-shot learning in Hatespeech task.}
\label{tab:prompt3}
\end{table}

\begin{table}[!ht]
\centering
\resizebox{\linewidth}{!}{
\begin{tabular}{p{8cm}|p{2.5cm}}
\hline
\textbf{Category} & \textbf{Threshold} \\ \hline
HARM\_CATEGORY\_HARASSMENT & BLOCK\_NONE \\
HARM\_CATEGORY\_HATE\_SPEECH & BLOCK\_NONE \\
HARM\_CATEGORY\_SEXUALLY\_EXPLICIT & BLOCK\_NONE \\
HARM\_CATEGORY\_DANGEROUS\_CONTENT & BLOCK\_NONE \\
HARM\_CATEGORY\_SEXUAL & BLOCK\_NONE \\
HARM\_CATEGORY\_DANGEROUS & BLOCK\_NONE \\
\bottomrule
\end{tabular}
}
\caption{Safety setting used for Gemini Pro model to prevent blocking the predictions for harmful content.}
\label{tab:safety_setting}
\end{table}
\section{Experimental Details and Results}
\label{sec:appendix_b}

\subsection{Experimental Settings}
\subsubsection{Data}
%\section{Data}
%\label{Data}

This section discusses the publicly available data for three tasks used in our study. We first discuss the data for the NLI task followed by the sentiment task and conclude with the hate speech task. Although each task has some datasets for all the languages individually, only the dataset of the NLI task has been translated into several languages. To fairly evaluate the generalization of LLMs, the translated version of the datasets is mandatory for other tasks. We provide a detailed description of data distribution in Table \ref{tab:data_distribution}.

\paragraph{NLI Task:}
We used the cross-lingual natural language inference (XNLI) dataset \cite{conneau2018xnli} for the NLI task. 
%The dataset extends the Multi-genre NLI dataset incorporating the raw text from the second release of the Open American National Corpus. The XNLI dataset is mainly developed for the English language and translated into 15 different languages including Hindi and Urdu languages using human annotators. Each data consists of a premise and hypothesis with a corresponding label\footnote{The class labels for the XNLI dataset are Contradiction, Entailment, and Neutral.}. During the development of the dataset, three different hypotheses were generated by the annotators based on the labels from each premise. 
We select the test set of English, Hindi, and Urdu languages from the XNLI dataset for our experiments.
For the Bangla language, we used the translated version of XNLI \cite{bhattacharjee2021banglabert}. %The dataset is translated using the English to Bangla translator model described in \cite{hasan-etal-2020-low}. 
%Although the dataset is translated from the XNLI dataset, the test set is short of 115 data from the original set. 

\paragraph{Sentiment Task:}

For the sentiment analysis task, we used the official test of SemEval-2017 task 4: Sentiment Analysis in Twitter \cite{SemEval:2017:task4}. 
%The raw texts were collected from X (formerly known as Twitter) and manually annotated them. 
Primarily, the annotation was completed in five classes %which include Strongly Positive, Positive, Neutral, Negative, and Strongly Negative. Later, 
and then the labels were re-mapped into three classes.% where Strongly Positive was combined with Positive and Strongly Negative with Negative classes. 
The SemEval-2017 task 4 offered only English and Arabic data. In this study, we only incorporate the English data.
%We translated the English test set for the Bangla, Hindi, and Urdu languages for evaluating the LLMs for the sentiment task. We used the web version of Google Translator\footnote{\url{https://translate.google.com}} with the use of Deep Translator toolkit\footnote{\url{https://pypi.org/project/deep-translator/}}. The quality of translations is moderate due to the tweet texts. We analyzed the translations and found that most of the hashtags were not translated into the target language. Moreover, Hindi translations were far better than Bangla and Urdu.

\paragraph{Hate Speech Task:}

We used the dataset described in \cite{hateoffensive} for our hate speech task. 
%The texts were collected from X (formerly known as Twitter) where the annotations were done manually into three categories that include 'Hate', 'Offensive', and 'Neither'. Each data was annotated by at least three people and the final label was consolidated by the majority. Few of the data were discarded where there was no majority. 
The official dataset consists of a total of $24,802$ samples. We first split the data into train, validation, and test splits by $70\%$, $10\%$, and $20\%$ respectively. We only used the test set in our study and the language of the official dataset is English.

\paragraph{Translation:}
We translated the English test set for the Bangla, Hindi, and Urdu languages to evaluate the LLMs for sentiment and hate speech tasks. We used the web version of Google Translator\footnote{\url{https://translate.google.com}} with the use of Deep Translator toolkit\footnote{\url{https://pypi.org/project/deep-translator/}}.
%The quality of translations is moderate due to the tweet texts. 
We analyzed the translations and found that most of the hashtags were not translated into the target language. Moreover, Hindi translations were far better than Bangla and Urdu. We also randomly sampled 100 translation pairs for each language from both tasks to check the translation quality by native speakers. The feedback from native speakers indicates that there is room for improvement in the translation quality. Additionally, it is important to note that we followed previous best practices used in similar studies \cite{aggarwal2022indicxnli, lai2023okapi}.

%For Bangla, Hindi, and Urdu language datasets, we translated the English test set using Google Translator. We randomly sampled 100 entries from each translated dataset to conduct an analysis on the quality of the translation. Our assessment revealed that while efforts were made, the quality of translation was found to be moderate, indicating room for improvement. Notably, certain elements such as hashtag words remained untranslated, and specific terms like 'hairspray,' 'oz,' numerical values, among others, were not adequately translated into their respective languages. 

\begin{table}[!ht]
\centering
\resizebox{\linewidth}{!}{
\begin{tabular}{lllr}
\toprule
\multicolumn{1}{c}{\textbf{Task}} & \multicolumn{1}{c}{\textbf{Languages}} & \multicolumn{1}{c}{\textbf{Class}} & \multicolumn{1}{c}{\textbf{Test}}\\ \midrule

\multirow{6}{*}{\textbf{NLI}} & \multirow{3}{*}{\textbf{EN, HI, UR}} & Contradiction & $1,670$ \\
& & Entailment & $1,670$ \\
& & Neutral & $1,670$ \\ \cline{2-4}
& \multirow{3}{*}{\textbf{BN}} & Contradiction & $1,630$ \\
& & Entailment & $1,631$ \\
& & Neutral & $1,634$ \\ \midrule
\multirow{3}{*}{\textbf{Sentiment}} & \multirow{3}{*}{\textbf{EN, BN, HI, UR}} & Negative & $3,972$ \\
& & Neutral & $5,937$ \\
& & Positive & $2,375$ \\ \midrule

\multirow{3}{*}{\textbf{Hate Speech}} & \multirow{3}{*}{\textbf{EN, BN, HI, UR}} & Hate & $280$ \\
& & Neither & $821$ \\
& & Offensive & $3,856$ \\
\bottomrule
\end{tabular}
}
\caption{Class-wise test set data distribution for all the tasks. EN: English, BN: Bangla, HI: Hindi, and UR: Urdu.}
\label{tab:data_distribution}
\end{table}

% \textcolor{blue}{We sampled 100 entries randomly from each translated dataset to analyze the translation quality. The quality of the translation is not good and the hashtags words were not translated. Moreover, some words such as `hairspray', `oz', numbers, etc. have also not been translated into their respective languages.}

%We also noticed the translation quality is moderate and has similar issues to the translation of sentiment task datasets.

\subsubsection{Data Pre-processing}
The sentiment and hate speech datasets were mainly collected from X and contain URLs, usernames, hashtags, emoticons, and symbols. We only removed the URLs and usernames from the sentiment and hate speech task datasets. We keep the hashtags, emoticons, and symbols with data to understand how LLMs performed with this mixed information. Moreover, we did not perform any preprocessing steps for the XNLI dataset.

\subsubsection{Evaluation Metrics}
To evaluate our experiments, we calculated accuracy, precision, recall, and F$_1$ scores for all the tasks. We computed the weighted version of precision and recall and the macro version of F$_1$ score as it considers class imbalance.

\subsection{Detailed Results}

We investigated the detailed performances of each task (see Table \ref{tab:nli_appendix}, Table \ref{tab:senti_appendix}, and Table \ref{tab:hate_appendix}). GPT-4 shows superior performances on the NLI task for all languages while exhibiting good performances on the sentiment task. However, most hate class data were misclassified in the hate speech task for all languages. Llama 2 provides strong performances in English for NLI, sentiment, and hate speech tasks while finding difficulties in accurately predicting the contradiction, neutral, and hate classes for NLI, sentiment, and hate speech tasks, respectively. Although Llama 2 outperforms GPT-4 performances in hate class in every language, GPT-4 in English and Hindi is better than Llama 2 for hate speech tasks. Moreover, Llama 2 demonstrated comparatively better performance on the hate speech task than NLI and sentiment tasks. While Gemini exhibits strong performances in NLI and sentiment tasks for all the languages, it consistently performs poorly on the speech task for all the languages. However, Gemini performs comparatively better hate class performance than Llama 2 and GPT-4 for all the languages. Moreover, the performances in the neither and offensive classes are worse than other LLMs. We also found that most offensive classes are misclassified as neither.

\subsubsection{NLI Task}

We present the detailed class-wise performances for the NLI task across the LLMs in Table \ref{tab:nli_appendix}.

\begin{table}[!ht]
\centering
\resizebox{\linewidth}{!}{
\begin{tabular}{lllrrr}
\toprule
\multicolumn{1}{c}{\textbf{Model}} & \multicolumn{1}{c}{\textbf{Lang.}} & \multicolumn{1}{c}{\textbf{Class}} & \multicolumn{1}{c}{\textbf{P.}} & \multicolumn{1}{c}{\textbf{R.}} & \multicolumn{1}{c}{\textbf{F1}}\\ \midrule

\multirow{12}{*}{\textbf{GPT-4}} & \multirow{3}{*}{\textbf{EN}} & Contradiction &  92.45 & 89.40 & 90.90\\
& & Entailment & 88.25 & 86.88 & 87.56 \\ 
& & Neutral &  80.02 & 82.90 & 81.92\\ \cline{2-6}

& \multirow{3}{*}{\textbf{BN}} & Contradiction & 85.58 & 67.03 & 75.18 \\
& & Entailment & 88.26 & 49.85 & 63.17 \\ 
& & Neutral & 54.10 & 89.24 & 67.36 \\ \cline{2-6}

& \multirow{3}{*}{\textbf{HI}} & Contradiction & 88.54 & 68.92 & 77.51\\
& & Entailment & 86.02 & 50.18 & 63.39 \\ 
& & Neutral & 54.22 & 88.80 & 67.33 \\ \cline{2-6}

& \multirow{3}{*}{\textbf{UR}} & Contradiction & 85.41 & 40.66 & 55.09\\
& & Entailment & 82.53 & 64.27 & 72.26\\ 
& & Neutral & 50.79 & 88.62 & 64.57\\ \midrule

\multirow{12}{*}{\textbf{Llama 2}} & \multirow{3}{*}{\textbf{EN}} & Contradiction &  94.12 & 73.83 & 82.75\\
& & Entailment & 72.88 & 83.17 & 77.68 \\ 
& & Neutral & 61.82 & 66.41 & 64.03\\ \cline{2-6}

& \multirow{3}{*}{\textbf{BN}} & Contradiction & 65.80 & 13.93 & 22.99\\
& & Entailment & 54.66 & 57.20 & 55.90 \\ 
& & Neutral & 37.81 & 65.79 & 48.02\\ \cline{2-6}

& \multirow{3}{*}{\textbf{HI}} & Contradiction & 88.30 & 14.91 & 25.51\\
& & Entailment & 70.72 & 41.80 & 52.54\\ 
& & Neutral & 38.01 & 85.15 & 52.56 \\ \cline{2-6}

& \multirow{3}{*}{\textbf{UR}} & Contradiction & 63.88 & 22.87 & 33.69\\
& & Entailment & 59.63 & 46.17 & 52.04\\ 
& & Neutral & 37.54 & 70.12 & 48.90\\ \midrule

\multirow{12}{*}{\textbf{Gemini}} & \multirow{3}{*}{\textbf{EN}} & Contradiction & 84.24 & 90.24 & 87.14 \\
& & Entailment & 77.76 & 80.00 & 78.87 \\ 
& & Neutral & 72.17 & 64.95 & 68.37\\ \cline{2-6}

& \multirow{3}{*}{\textbf{BN}} & Contradiction & 72.90 & 78.81 & 75.57 \\
& & Entailment & 79.22 & 53.35 & 63.76 \\ 
& & Neutral & 55.88 & 69.57 & 61.97\\ \cline{2-6}

& \multirow{3}{*}{\textbf{HI}} & Contradiction & 74.14 & 75.36 &74.73\\
& & Entailment & 77.08 & 53.21 &62.96\\ 
& & Neutral & 54.82 & 70.88 &61.82\\ \cline{2-6}

& \multirow{3}{*}{\textbf{UR}} & Contradiction & 70.14 & 70.06 &70.10\\
& & Entailment & 75.27 & 45.81 &56.98\\ 
& & Neutral & 50.62 & 70.54 &58.94\\ \midrule

\bottomrule
\end{tabular}
}
\caption{Class-wise performances of the NLI task across the models and languages. \textbf{Bold} indicates the best performances across the languages. Lang.: language, P.: Precision, R.: Recall, EN: English, BN: Bangla, HI: Hindi, and UR: Urdu}
\label{tab:nli_appendix}
\end{table}

\subsubsection{Sentiment Task}

Detailed class-wise performances for the sentiment task across the LLMs are presented in Table \ref{tab:senti_appendix}.

\begin{table}[!ht]
\centering
\resizebox{\linewidth}{!}{
\begin{tabular}{lllrrr}
\toprule
\multicolumn{1}{c}{\textbf{Model}} & \multicolumn{1}{c}{\textbf{Lang.}} & \multicolumn{1}{c}{\textbf{Class}} & \multicolumn{1}{c}{\textbf{P.}} & \multicolumn{1}{c}{\textbf{R.}} & \multicolumn{1}{c}{\textbf{F1}}\\ \midrule

\multirow{12}{*}{\textbf{GPT-4}} & \multirow{3}{*}{\textbf{EN}} & Negative & 73.08 & 73.39 &73.23 \\
& & Neutral & 70.52 & 77.23 & 73.72\\ 
& & Positive & 79.36 & 59.92 & 68.28\\ \cline{2-6}

& \multirow{3}{*}{\textbf{BN}} & Negative & 71.29 & 39.88 & 51.15\\
& & Neutral & 57.40 & 85.11 & 68.56 \\ 
& & Positive & 71.25 & 37.77 & 49.37\\ \cline{2-6}

& \multirow{3}{*}{\textbf{HI}} & Negative & 73.07 & 51.79 & 60.62\\
& & Neutral & 62.03	& 83.90 & 71.33\\ 
& & Positive & 78.32 & 47.45 & 59.10 \\ \cline{2-6}

& \multirow{3}{*}{\textbf{UR}} & Negative & 72.34 & 43.01 & 53.95\\
& & Neutral & 58.45	& 83.43	& 68.74\\ 
& & Positive & 68.51 & 41.77 & 51.90\\ \midrule

\multirow{12}{*}{\textbf{Llama 2}} & \multirow{3}{*}{\textbf{EN}} & Negative & 56.08	& 94.26	& 70.32 \\
& & Neutral & 81.81	& 16.89	& 28.01 \\ 
& & Positive & 47.65 & 87.92 & 61.80 \\ \cline{2-6}

& \multirow{3}{*}{\textbf{BN}} & Negative & 45.10 & 90.79 & 60.27 \\
& & Neutral & 76.96	& 2.81 & 5.43 \\ 
& & Positive & 43.66 & 74.89 & 55.16\\ \cline{2-6}

& \multirow{3}{*}{\textbf{HI}} & Negative & 48.31 & 93.78 & 63.77\\
& & Neutral & 80.45	& 4.78 & 9.03\\ 
& & Positive & 45.62 & 81.05 & 58.38 \\ \cline{2-6}

& \multirow{3}{*}{\textbf{UR}} & Negative & 46.15 & 93.55 & 61.81\\
& & Neutral & 78.18	& 4.77 & 8.99\\ 
& & Positive & 46.05 & 75.03 & 57.07\\ \midrule

\multirow{12}{*}{\textbf{Gemini}} & \multirow{3}{*}{\textbf{EN}} & Negative & 60.40 & 87.89	& 71.60 \\
& & Neutral & 76.83	& 46.38	& 57.84 \\ 
& & Positive & 57.86 & 71.33 & 63.89 \\ \cline{2-6}

& \multirow{3}{*}{\textbf{BN}} & Negative & 61.28 & 84.21 & 70.94 \\
& & Neutral & 72.07	& 54.44	& 62.03 \\ 
& & Positive & 62.23 & 61.42 & 61.82\\ \cline{2-6}

& \multirow{3}{*}{\textbf{HI}} & Negative & 62.57 & 83.42 & 71.51\\
& & Neutral & 71.36	& 57.17	& 63.48\\ 
& & Positive & 62.33 & 58.65 & 60.43 \\ \cline{2-6}

& \multirow{3}{*}{\textbf{UR}} & Negative & 61.74 & 84.66 & 71.41\\
& & Neutral & 72.63	& 55.11	& 62.67\\ 
& & Positive & 62.41 & 61.42 & 61.91\\ \midrule

\bottomrule
\end{tabular}
}
\caption{Class-wise performances of the Sentiment task across the models and languages. \textbf{Bold} indicates the best performances across the languages. Lang.: language, P.: Precision, R.: Recall, EN: English, BN: Bangla, HI: Hindi, and UR: Urdu}
\label{tab:senti_appendix}
\end{table}

\subsubsection{Hatespeech Task}
Table \ref{tab:hate_appendix} reports the detailed class-wise performances for the hatespeech task across the LLMs.

\begin{table}[!ht]
\centering
\resizebox{\linewidth}{!}{
\begin{tabular}{lllrrr}
\toprule
\multicolumn{1}{c}{\textbf{Model}} & \multicolumn{1}{c}{\textbf{Lang.}} & \multicolumn{1}{c}{\textbf{Class}} & \multicolumn{1}{c}{\textbf{P.}} & \multicolumn{1}{c}{\textbf{R.}} & \multicolumn{1}{c}{\textbf{F1}}\\ \midrule

\multirow{12}{*}{\textbf{GPT-4}} & \multirow{3}{*}{\textbf{EN}} & Hate & 62.96 & 12.14 & 20.36 \\
& & Offensive & 88.85 & 95.10 & 91.87 \\ 
& & Neither & 77.58	& 73.33	& 75.39 \\ \cline{2-6}

& \multirow{3}{*}{\textbf{BN}} & Hate & 22.39 & 5.36 & 8.65 \\
& & Offensive & 89.56 & 51.61 & 65.48 \\ 
& & Neither & 27.62	& 89.77	& 42.25\\ \cline{2-6}

& \multirow{3}{*}{\textbf{HI}} & Hate & 32.69 & 6.07 & 10.24\\
& & Offensive & 90.97 & 63.49 & 74.68\\ 
& & Neither & 33.56 & 90.13	& 48.91 \\ \cline{2-6}

& \multirow{3}{*}{\textbf{UR}} & Hate & 33.93 & 6.79 & 11.31\\
& & Offensive & 88.58 & 50.49 & 64.32\\ 
& & Neither & 26.30	& 86.60	& 40.35 \\ \midrule

\multirow{12}{*}{\textbf{Llama 2}} & \multirow{3}{*}{\textbf{EN}} & Hate & 14.98 & 31.79 & 20.37 \\
& & Offensive & 88.16 & 86.51 & 87.33 \\ 
& & Neither & 87.56	& 61.75	& 72.43 \\ \cline{2-6}

& \multirow{3}{*}{\textbf{BN}} & Hate & 13.35 &	17.50 &	15.15 \\
& & Offensive & 80.82 &	85.14 &	82.92 \\ 
& & Neither & 42.42 & 27.28 & 33.21\\ \cline{2-6}

& \multirow{3}{*}{\textbf{HI}} & Hate & 15.09 & 12.50 &	13.67\\
& & Offensive & 80.93 &	89.06 &	84.80\\ 
& & Neither & 46.89	& 27.53	& 34.69 \\ \cline{2-6}

& \multirow{3}{*}{\textbf{UR}} & Hate & 11.98 & 18.57 &	14.57\\
& & Offensive & 80.05 & 83.87 &	81.91\\ 
& & Neither & 37.27	& 21.92	& 27.61\\ \midrule

\multirow{12}{*}{\textbf{Gemini}} & \multirow{3}{*}{\textbf{EN}} & Hate & 14.95 &	76.34 &	25.00 \\
& & Offensive & 88.87 &	55.49 &	68.32 \\ 
& & Neither & 46.97 & 63.41	& 53.97 \\ \cline{2-6}

& \multirow{3}{*}{\textbf{BN}} & Hate & 8.62 & 79.93 & 15.56 \\
& & Offensive & 83.14 & 20.36 & 32.71 \\ 
& & Neither & 34.83	& 60.29	& 44.16\\ \cline{2-6}

& \multirow{3}{*}{\textbf{HI}} & Hate & 8.27 & 81.65 & 15.01\\
& & Offensive & 83.90 & 22.50 & 35.49\\ 
& & Neither & 42.47	& 59.51	& 49.57 \\ \cline{2-6}

& \multirow{3}{*}{\textbf{UR}} & Hate & 8.76 & 76.43 & 15.72\\
& & Offensive & 83.20 & 18.53 & 30.31\\ 
& & Neither & 29.49	& 59.20	& 39.37\\

\bottomrule
\end{tabular}
}
\caption{Class-wise performances of the Hatespeech task across the models and languages. \textbf{Bold} indicates the best performances across the languages. Lang.: language, P.: Precision, R.: Recall, EN: English, BN: Bangla, HI: Hindi, and UR: Urdu}
\label{tab:hate_appendix}
\end{table}
\section{Experimental Analysis}

\end{document}